\documentclass[letterpaper, 10 pt, conference]{ieeeconf}
\IEEEoverridecommandlockouts
\usepackage{caption}
\usepackage{hyperref}
\usepackage{graphicx}
\usepackage{multirow}
\usepackage{booktabs}
\usepackage{pifont}
\usepackage{amssymb}
\usepackage{amsmath}
\usepackage{makecell}
\usepackage{rotating}
\usepackage{xcolor}

\title{\LARGE \bf
LightStereo: Channel Boost Is All You Need \\ for Efficient 2D Cost Aggregation
}

\author{
    Xianda Guo\textsuperscript{1,$^*$},
    Chenming Zhang\textsuperscript{2,3,$^*$},
    Youmin Zhang\textsuperscript{4,5},
Wenzhao Zheng\textsuperscript{6},\\
Dujun Nie\textsuperscript{7,8},
Matteo Poggi\textsuperscript{4},
    Long Chen\textsuperscript{7,2,3,$^\dagger$} \\
    \textsuperscript{1} School of Computer Science, Wuhan University
    \textsuperscript{2} IAIR, Xi'an Jiaotong University\\
~~~~\textsuperscript{3} Waytous ~~~~\textsuperscript{4} University of Bologna ~~~~\textsuperscript{5} Rock Universe AI ~~~~\textsuperscript{6}University of California, Berkeley  \\ ~~~~\textsuperscript{7} Institute of Automation, Chinese Academy of Sciences ~~~~\textsuperscript{8} Metoak
\\
\texttt{xianda\_guo@163.com; 
chmzhang@outlook.com;
long.chen@ia.ac.cn}
\vspace{-5mm}
}

\begin{document}
\maketitle

\renewcommand{\thefootnote}{\fnsymbol{footnote}}
\footnotetext[1]{These authors contributed equally to this work.}
\footnotetext[2]{Corresponding Author}

\begin{abstract}

We present LightStereo, a cutting-edge stereo-matching network crafted to accelerate the matching process. Departing from conventional methodologies that rely on aggregating computationally intensive 4D costs, LightStereo adopts the 3D cost volume as a lightweight alternative. While similar approaches have been explored previously, our breakthrough lies in enhancing performance through a dedicated focus on the channel dimension of the 3D cost volume, where the distribution of matching costs is encapsulated. Our exhaustive exploration has yielded plenty of strategies to amplify the capacity of the pivotal dimension, ensuring both precision and efficiency. We compare the proposed LightStereo with existing state-of-the-art methods across various benchmarks, which demonstrate its superior performance in speed, accuracy, and resource utilization. LightStereo achieves a competitive EPE metric in the SceneFlow datasets while demanding a minimum of only 22 GFLOPs and 17 ms of runtime, and ranks 1st on KITTI 2015 among real-time models.
Our comprehensive analysis reveals the effect of 2D cost aggregation for stereo matching, paving the way for real-world applications of efficient stereo systems.
Code is available at \url{https://github.com/XiandaGuo/OpenStereo}.

\end{abstract}
\section{Introduction}

Stereo matching is a pivotal task in computer vision, which aims to ascertain correspondences between points in stereo image pairs to retrieve depth information. This process underpins numerous applications, including autonomous driving, robotic navigation, and augmented reality. Despite substantial advancements, achieving real-time stereo matching without sacrificing accuracy or efficiency remains a formidable challenge, especially on 
embedded platforms.

Most developments~\cite{gcnet, psmnet2018, gwcnet2019, ganet2019, dsmnet2020, gu2020cascade,leastereo, wang2021pvstereo,song2021adastereo, Crestereo,liu2022local,acvnet,nie2019multi,guo2023openstereo} have primarily focused on leveraging deep learning techniques for accurate stereo matching. Most methods are based on 3D CNN for cost aggregation: 
because of the disparity dimension being modeled in the constructed 4D cost volume, the network can perform exhaustively aggregation for accurate matching, yet at the expense of high memory consumption and runtime. 
Nevertheless, the overall runtime of these iterative-based methods is over 100ms on a custom GPU.

\begin{figure}[t]
      \centering
      \includegraphics[width=0.98\linewidth]{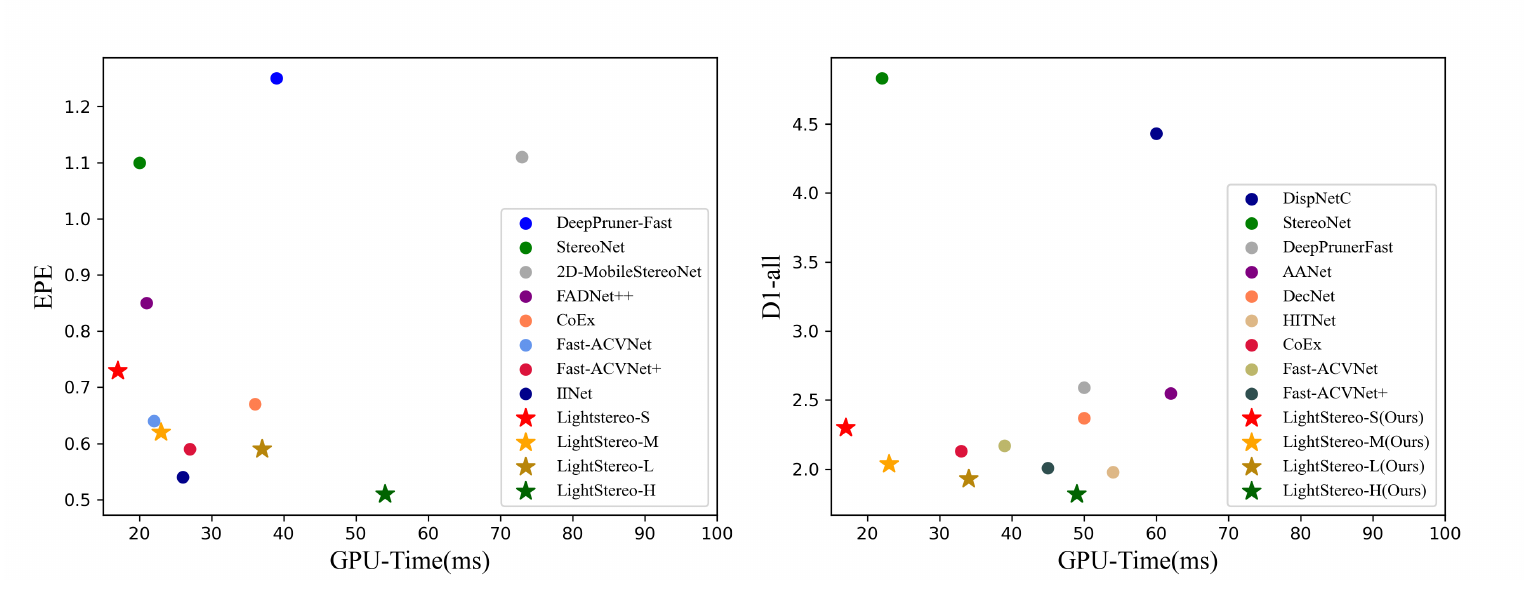}
      \caption{\textbf{Performence vs. Time trade-off.} Results on SceneFlow (Left) and KITTI15 (Right) datasets.} 
      \vspace{-0.5cm}
      \label{fig:GPU_time}
\end{figure}

\begin{figure*}[t]
      \centering
      \includegraphics[width=0.95\linewidth]{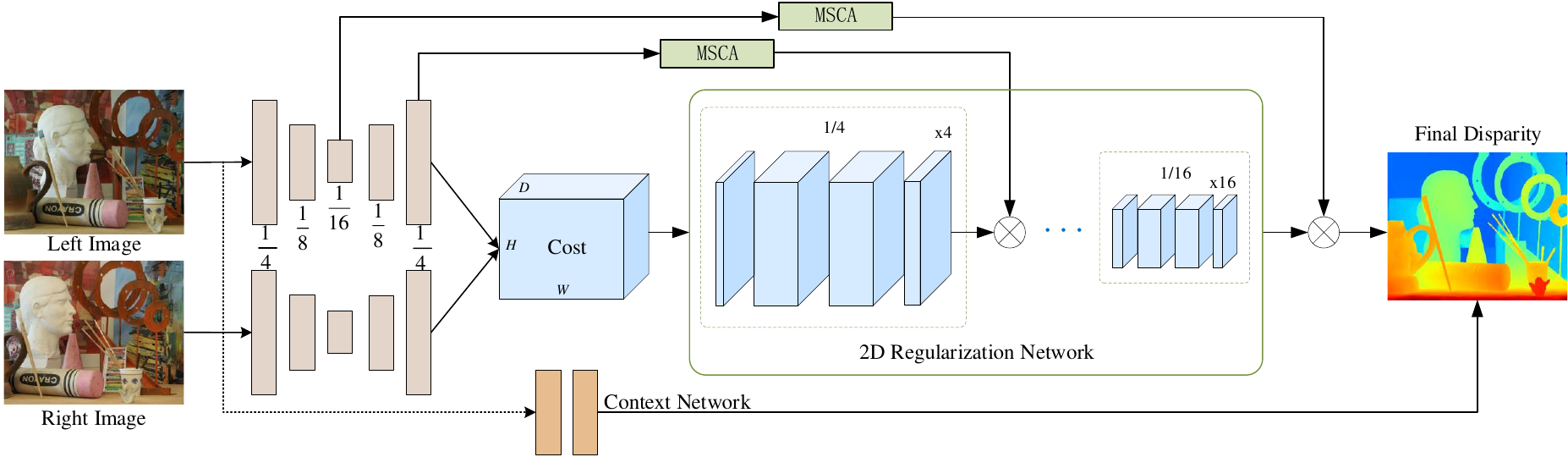}
      \caption{\textbf{The diagram above illustrates the architecture of LightStereo. }} 
      \vspace{-0.65cm}
      \label{fig:lightstereo}
\end{figure*}

There are also efforts~\cite{wang2020fadnet,wang2021fadnet++, shamsafar2022mobilestereonet,xu2020aanet, xu2023cgi} focusing on lightweight design for stereo matching.
AANet~\cite{xu2020aanet} constructs a 3D cost volume by correlating the left and right images and introduces intra-scale and cross-scale 2D cost aggregation modules to enhance the efficiency and accuracy of cost aggregation. MobilenetStereo-2D~\cite{shamsafar2022mobilestereonet} introduces MobileNet~\cite{mijwil2023mobilenetv1, sandler2018mobilenetv2} blocks to reduce the cost, but the performance is far from satisfaction. Overall, these methods based on 2D cost aggregation perform poorly. 
Cost aggregation is critical to accuracy and efficiency in stereo matching yet existing methods require a compromise between accuracy and speed. However, existing methodologies often necessitate a trade-off between accuracy and processing speed. This paper poses the question: Is it possible to design a lightweight 2D encoder-decoder aggregation net to achieve precise disparity estimation?

From the perspective of cost aggregation in stereo matching, focusing on the disparity channel dimension presents several advantages. Firstly, it allows for more direct modeling of the disparities between corresponding image points, which is crucial for accurate disparity estimation. By focusing on this dimension, it becomes possible to more effectively capture and assimilate the critical information needed for robust stereo matching.
In this paper, we propose LightStereo as a positive solution.
We explore 2D cost aggregation for stereo matching and leverage inverted residual blocks to enhance both accuracy and computational efficiency. 
The model is specifically designed to address the challenges of real-time stereo vision applications, focusing on reducing computational demands without compromising the quality of disparity estimation. Specifically, we utilize inverted residual blocks for 2D cost aggregation in stereo matching, which focuses on the disparity channel dimension of the 3D cost volume rather than the dimension of height and width. 
In inverted residual blocks, the expansion phase increases the channel dimension, allowing the network to learn richer features at a reduced computational cost before compressing them back. 
In addition, inspired by the effectiveness of large kernel and strip convolutions in image segmentation~\cite{peng2017large,hou2020strip,guo2022segnext}, we propose a Multi-Scale Convolutional Attention Module (MSCA) module for enhancing cost aggregation by extracting features from the left image. By leveraging multi-scale image features to excite the channel dimension of the cost volume, we utilize semantic information inherent in the images (such as object-level semantic details) to guide the cost aggregation process. When encountering discontinuities in disparity, the network halts propagation.

Our main contributions are as follows: 
\begin{itemize}

\item We propose LightStereo, which achieves a competitive End-Point-Error (EPE) in the SceneFLow~\cite{sceneflow} datasets while demanding a minimum of only 22 Gflops with an inference time of 17 ms.

\item We propose inverted residual blocks for 2D cost aggregation in stereo matching.

\item We propose the MSCA module for enhancing cost aggregation by extracting features from the left image.

\item We verify the effectiveness of LightStereo, achieving state-of-the-art performance on the Sceneflow~\cite{sceneflow} and KITTI~\cite{kitti2012,kitti2015} benchmark within the lightweight stereo-matching methods.
\end{itemize}

\section{Related Work}

We review the literature concerning deep learning-based stereo matching, identifying two categories of methods. We refer the reader to \cite{poggi2021synergies,tosi2024surveydeepstereomatching} for a broader overview.

\noindent \textbf{Accuracy-Focused Stereo Matching.}
Many contemporary stereo-matching methods are dedicated to enhancing accuracy through various techniques and optimizations. Among these, 3D end-to-end networks have introduced significantly heightened precision in disparity estimation within stereo matching~\cite{psmnet2018, gwcnet2019, leastereo, acvnet, xu2023iterative, guo2023openstereo}. PSMNet~\cite{psmnet2018} adopts an architecture that leverages spatial pyramid pooling and 3D convolutional neural networks to learn disparity estimation from stereo images. GwcNet~\cite{gwcnet2019} introduces a novel approach for constructing the cost volume in stereo matching using group-wise correlation. By dividing left and right features into groups along the channel dimension, correlation maps are computed within each group to generate multiple matching cost proposals. LEAStereo~\cite{leastereo} employs a hierarchical neural architecture search (NAS) framework to enhance deep stereo matching performance.
IGEV~\cite{xu2023iterative} constructs a unified geometry encoding volume, integrating geometry, contextual cues, and local matching intricacies. 
OpenStereo~\cite{guo2023openstereo} conducted a comprehensive benchmark with a focus on practical applicability and introduced StereoBase, which further elevates the performance ceiling of stereo matching.
Selective-IGEV~\cite{SelectiveStereo} proposes the Selective Recurrent Unit to help integrate disparity information across frequencies, minimizing loss during iterative steps.
In addition, work ~\cite{ganetADL} improves the loss function by introducing ADL, an adaptive multi-modal cross-entropy loss, to guide network learning of diverse pixel distribution patterns.

\noindent \textbf{Real-Time-Focused Stereo Matching.}
A multitude of models have been proposed to target practical applications.
StereoNet~\cite{khamis2018stereonet} uses color inputs to guide hierarchical refinement and can recover high-frequency details. DeepPruner\cite{deeppruner2019} proposes a PatchMatch module with learnable parameters to save memory and computation by gradually trimming the disparity space to be searched for each pixel.  AnyNet\cite{anynet2019} has a 2D image convolutional network and a tiny 3D cost aggregation network. 
FADNet\cite{wang2020fadnet} leverages efficient 2D correlation layers with residual structures to perform multi-scale predictions. Based on the learned bilateral grid, BGNet\cite{bgnet} designs an edge-preserving cost volume up-sampling module, allowing computationally expensive operations such as 3D convolution to be performed at low resolution. MADNet series \cite{tonioni2019real,poggi2024federated} relies on fully 2D modules, maximizing speed at the expense of accuracy. CoEx\cite{bangunharcana2021coex} uses image feature-guided weights and cost volume to excite 3D CNN to extract relevant geometric features. 
Building upon the foundation of StereoNet~\cite{khamis2018stereonet}, MobileStereoNet\cite{shamsafar2022mobilestereonet} introduces two models suitable for resource-constrained devices. 

However, these methods still have considerable room for improvement in terms of computational efficiency and parameter scale. 
After carefully considering and testing various network structures, we propose a lightweight network, called LightStereo. This model innovatively explores 2D cost aggregation for stereo matching, leveraging advanced techniques to enhance both accuracy and computational efficiency.

\begin{figure}[t]
      \centering
      \includegraphics[width=0.9\linewidth]{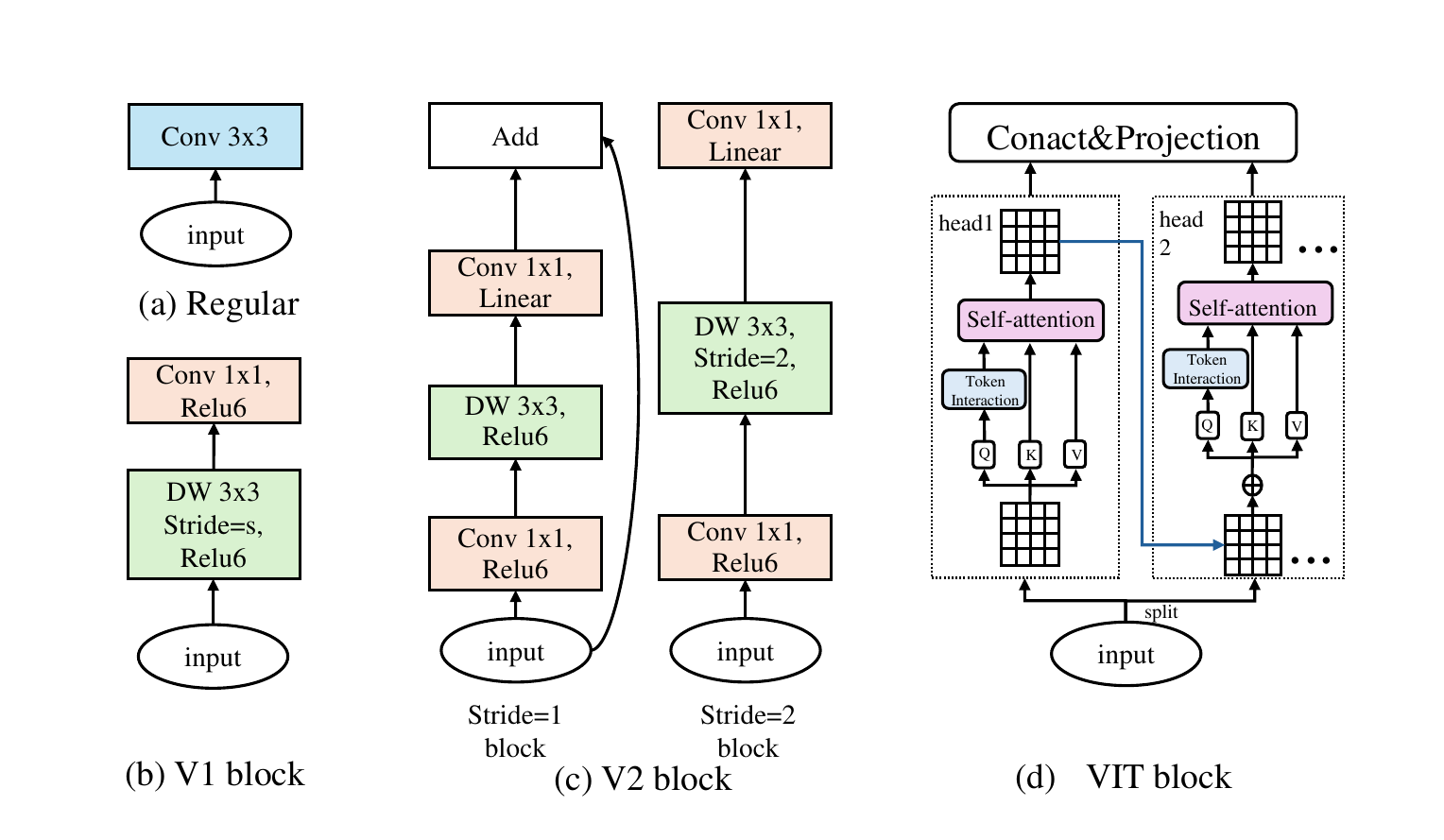}
      \vspace{-0.4cm}
      \caption{\textbf{Comparison of different blocks used for cost aggregation} -- DW refers to depthwise separable convolution; V1 Block represents the depthwise separable convolution~\cite{mijwil2023mobilenetv1}; V2 Block represents an inverted residual block~\cite{sandler2018mobilenetv2}; ViT Block refers to the block used in EfficientViT~\cite{liu2023efficientvit}.} 
      \label{fig:blockComprasion}
      \vspace{-0.4cm}
\end{figure}
\section{Method} \label{method}

In the deployment of stereo matching on edge devices, constructing a 4D cost volume and using 3D CNNs for cost aggregation prove to be extremely inefficient. Our objective is to construct a 3D cost volume, where single scores are stored along the channel dimension, and utilize 2D CNNs augmented with channel boost for cost aggregation to balance efficiency and accuracy. 
We are now going to introduce LightStereo, whose overview is shown in Figure~\ref{fig:lightstereo}, starting with its main components -- the Inverted Residual Blocks \cite{sandler2018mobilenetv2} and the new Multi-Scale Convolutional Attention -- and concluding with the overall architectural design.

\subsection{Inverted Residual Blocks for 2D Cost Aggregation}  
\label{InvertedResidualblocks}

Previous efforts, such as MobilenetStereo-2D~\cite{shamsafar2022mobilestereonet}, have incorporated MobileNet~\cite{mijwil2023mobilenetv1,sandler2018mobilenetv2} blocks to decrease computational costs. Despite these efforts, the results have not met expectations, with an EPE of 1.14 on SceneFLow~\cite{sceneflow} still being reported.
To address this issue, our study introduces inverted residual blocks to boost disparity estimation accuracy, as shown in Figure~\ref{fig:blockComprasion} (c). 
Given the cost volume in Figure~\ref{fig:lightstereo} whose size is $\frac{H}{4}\times\frac{W}{4}\times\frac{Disp}{4}$, the key idea is first to boost the number of disparity channels, then apply depthwise convolution, and finally project the expanded features back to a lower-dimensional space, enhancing feature representation significantly. 
Inverted residual blocks are utilized at resolutions of $\frac{1}{4}$, $\frac{1}{8}$, and $\frac{1}{16}$, each corresponding to different blocks: 
initially, the cost volume \( \mathbf{C} \) is first passed through a $1 \times 1$ convolution to increase the number of disparity channel:
\begin{equation}
\mathbf{y} = \sigma(\mathbf{W}_{\text{expand}} \ast \mathbf{C}),
\end{equation}
where \( \mathbf{W}_{\text{expand}} \) represents the weights of the expansion convolution, \( \ast \) denotes the convolution operation, and \( \sigma \) is the ReLU6 activation function.
Next, the expanded feature map \( \mathbf{y} \) undergoes a  $3 \times 3$  depthwise convolution, which operates independently on each channel to capture spatial features:
\begin{equation}
\mathbf{z} = \sigma(\mathbf{W}_{\text{depthwise}} \ast \mathbf{y}),
\end{equation}
where \( \mathbf{W}_{\text{depthwise}} \) are the weights of the depthwise convolution.
Finally, the result \( \mathbf{z} \) is then passed through another $1 \times 1$ convolution to restore the original number of channels: 
\begin{equation}
\mathbf{out} = \mathbf{W}_{\text{project}} \ast \mathbf{z},
\end{equation}
where \( \mathbf{W}_{\text{project}} \) represents the weights of the projection convolution.
If the input and output dimensions match, a skip connection is added to improve gradient flow: 
\begin{equation}
\mathbf{out} = \mathbf{out} + \mathbf{x} \quad \text{if dimensions match}.
\end{equation}

The structure of the inverted residual block significantly reduces computational complexity, making it ideal for resource-constrained environments. Our experimental results will demonstrate its superiority over regular CNN block, V1 block~\cite{mijwil2023mobilenetv1}, and Vision Transformer (ViT) block~\cite{liu2023efficientvit} -- depicted in Figure~\ref{fig:blockComprasion} (a), (b) and (d) respectively.

\begin{figure}[t]
      \centering
      \includegraphics[width=0.65\linewidth]{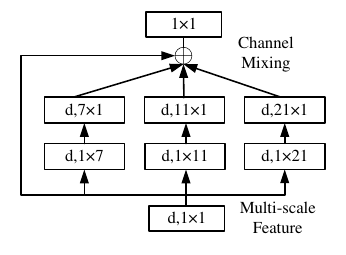}
      \vspace{-0.35cm}
      \caption{\textbf{Multi-Scale Convolutional Attention (MSCA).}} 
      \vspace{-0.35cm}
      \label{fig:msca}
\end{figure}

\subsection{Multi-Scale Convolutional Attention Module}
\label{MSCA}
Inspired by the effectiveness of large kernel and strip convolutions in image segmentation~\cite{peng2017large,hou2020strip,guo2022segnext}, we propose a Multi-Scale Convolutional Attention (MSCA) module for enhancing cost aggregation by extracting features from the left image. 
Figure~\ref{fig:msca} illustrates the architecture of such a module, designed to capture and integrate features at multiple scales and enhance their representation for cost aggregation. 
The MSCA incorporates a series of depthwise separable convolutions with varying kernel sizes, specifically \(1 \times 1\), \(7 \times 1\), \(1 \times 7\), \(11 \times 1\), \(1 \times 11\), \(21 \times 1\), and \(1 \times 21\), 
capturing both horizontal and vertical strip-like features, which are crucial for identifying elongated structures within the image. 
These depth-wise strip convolutions are lightweight and can replace, for instance, a standard $7 \times 7$ convolution with $7 \times 1$ and $1 \times 7$ convolutions.
Thus, strip convolutions serve as a complement to grid convolutions and aid in extracting strip-like features of left images. Utilizing multi-scale image features, MSCA enhances the channel dimension of the cost volume, 
by incorporating semantic information embedded within the images to guide the aggregation process. The network is designed to cease propagation upon detecting disparity discontinuities.

Given a stereo image of size $H\times W\times 3$, feature maps at $\frac{1}{4}$, $\frac{1}{8}$, and $\frac{1}{16}$ of the original resolution, and processed by the MSCA module to further extract horizontal and vertical strip-like features.
The outputs are then concatenated to form a comprehensive multi-scale feature representation, subsequently processed by a $1 \times 1$ convolution, which acts as a channel mixer. The $1 \times 1$ convolution blends the multi-scale features and recalibrates the feature channels, allowing the network to focus on relevant information across different scales. 
The final output is combined with the aggregated cost by multiplication, enhancing the effect of cost aggregation.

\subsection{LightStereo -- Network Architecture}
\label{lightstereo}

We now describe the overall LightStereo architecture, which consists of four components: feature extraction, cost computation, cost aggregation, and disparity prediction.

\textbf{Multi-scale Feature Extraction.}
For a stereo image pair with dimensions $H\times W\times 3$, our network harnesses a MobileNetV2~\cite{bangunharcana2021coex,xu2023iterative} model, previously trained on the ImageNet~\cite{deng2009imagenet} dataset.
We enable the extraction of feature maps across four distinct scales, effectively reducing the resolution to $\frac{1}{4}$, $\frac{1}{8}$, and $\frac{1}{16}$ and $\frac{1}{32}$ of the initial size, respectively. Subsequently, upsampling blocks with skip connections are utilized to restore these feature maps to $\frac{1}{4}$ scale, which is used to construct the cost volume.

\textbf{Cost Volume.} We construct a correlation volume from the previous left $\mathbf{f}_{l,4}$ and right features $\mathbf{f}_{r,4}$.
For each disparity \(d\) within the range of 0 to \(D-1\), the similarity between $\mathbf{f}_{l,4}$ and $\mathbf{f}_{r,4}$ shifted by \(d\) pixels is computed as: 

\begin{equation}
C_{cor}(d, h, w) = \frac{1}{C} \sum_{c=1}^{C} {f}_{l,4}(h, w) \cdot {f}_{r,4}(h, w-d).
\end{equation}

For \(d = 0\), the cost is computed as the average of the element-wise product of feature vectors from $\mathbf{f}_{l,4}$  and $\mathbf{f}_{r,4}$ at the same spatial location, across all channels.

\textbf{Cost Aggregation.} We use inverted residual blocks to aggregate the cost volume at $\frac{1}{4}$, $\frac{1}{8}$, and $\frac{1}{16}$ resolutions, and at each resolution, we apply multi-scale convolutional attention using the left image, as described in Section \ref{InvertedResidualblocks} and Section~\ref{MSCA}.
We develop four variants of LightStereo based on block and expansion variations: LightStereo-S for small-scale applications, LightStereo-M for medium-scale tasks, LightStereo-L for large-scale tasks, and LightStereo-H for huge-scale tasks.
In LightStereo-S, the inverted residual blocks are configured as (1, 2, 4) with an expansion factor of 4. For LightStereo-M, the blocks are set as (4, 8, 14) with an expansion factor of 4. Finally, LightStereo-L uses blocks (8, 16, 32) with an expansion factor of 8. LightStereo-H denotes a LightStereo-L variant with EfficientnetV2~\cite{tan2021efficientnetv2} backbone.

\textbf{Disparity Regression.} 
We use soft-argmax~\cite{gcnet,psmnet2018} to predict the final 
disparity map $\hat{d}$, determined by summing each disparity $d$ weighted by its probability $\sigma(c_d)$, with $c_d$ being the predicted cost and $\sigma(\cdot)$ a softmax layer:

\begin{equation}
\hat{d} = \sum_{d=0}^{D_{\text{max}}} d \times \sigma(c_d).
\end{equation}

\textbf{Training Loss.} We train LightStereo with smooth $L_1$ loss

\begin{equation}
L(d, \hat{d}) = \frac{1}{N} \sum_{i=1}^{N} \text{smooth}_{L_1}(d_i - \hat{d_i}),
\end{equation}

where $N$ is the number of labeled pixels, $d$ represents the ground-truth disparity, and $\hat{d}$ denotes the predicted disparity.

\begin{table}[t]
    \centering
    \caption{\textbf{Results on SceneFlow \cite{sceneflow}.} Comparison with state-of-the-art models pursuing speed over accuracy. 
    }
    \resizebox{0.48\textwidth}{!}{
    \begin{tabular}{l|cccc}
     \toprule
     Method& FLOPs(G) & Params(M) & EPE (px) & Time(ms) \\
     \midrule
     DeepPruner-Fast\cite{deeppruner2019} & 219.12 & 7.47 &1.25 &39 \\
     StereoNet\cite{stereonet2018} & 85.93 & \textbf{0.40} &1.10 &\underline{20} \\     
     2D-MobileStereoNet\cite{shamsafar2022mobilestereonet} & 128.84 & 2.23 &1.11 &73 \\
     FADNet++\cite{wang2021fadnet++} & 148.21 & 124.26 &0.85 &21 \\
     CoEx\cite{bangunharcana2021coex} & 53.39 & 2.72 &0.67 &36 \\
     Fast-ACVNet~\cite{fast-acv}&79.34&3.08&0.64&22 \\
     Fast-ACVNet+~\cite{fast-acv}&93.08&3.20&0.59&27 \\
     AANet\cite{xu2020aanet} &152.86  &2.97 &0.87 &93\\
      HITNet~\cite{hitnet}& 50.23&\underline{0.42}&0.55& 36\\ 
     IINet\cite{li2024iinet}&90.16& 19.54&\underline{0.54} &26 \\
     \textbf{LightStereo-S(Ours)} & \textbf{22.71}&3.44 &0.73 &\textbf{17}\\
     \textbf{LightStereo-M(Ours)} & \underline{36.36}&7.64 &0.62&23\\
     \textbf{LightStereo-L(Ours)} & 91.85&24.29&0.59 &37\\
    \textbf{LightStereo-H(Ours)} &159.26&45.63&\textbf{0.51}&54\\
    \bottomrule
    \end{tabular}
    }
\label{tab:scene_flow}
\vspace{-0.5cm}
\end{table}


\section{Experiments} \label{experiment}

\subsection{Datasets and Evaluation Metrics}

\textbf{SceneFlow}\cite{sceneflow} is a synthetic stereo collection that provides 35\,454 and 4\,370 image pairs for training and testing, respectively. The dataset has a resolution of 960$\times$540 and provides dense disparity maps as ground truth. Here, we use the End-Point-Error (EPE) as the evaluation metric.

\textbf{KITTI.} KITTI 2012 \cite{kitti2012} and 2015 \cite{kitti2015} are real-world datasets, counting 194/195 and 200/200 image pairs for training and testing, respectively, for which sparse ground-truth disparities are collected with a LiDAR system. 
We both report the results obtained from the online benchmarks, for which we report the official metrics, 
as well as assess cross-domain generalization performance, measuring the percentage of pixels with error \textgreater3px in this latter case.

\textbf{Middlebury 2014.}~\cite{middlebury} It collects high-resolution indoor scenes, providing 15 image pairs for training and 15 for testing, respectively. 
We involved this dataset for cross-domain generalization experiments, using half-resolution images and measuring the percentage of pixels with error \textgreater2px.

\subsection{Implementation Details}

LightStereo is implemented in PyTorch and trained on 8 NVIDIA RTX 3090 GPUs. 
For the SceneFlow~\cite{sceneflow} dataset, the batch sizes for LightStereo-S, LightStereo-M, LightStereo-L, and LightStereo-H are set at 24, 12, 8, and 6, respectively.We use the AdamW optimizer coupled with OneCycleLR scheduling, where the maximum learning rate was set to 0.0001 multiplied by the batch size. LightStereo underwent training for 90 epochs. Only random crop ($320 \times 736$) is employed.
For the KITTI dataset, we fine-tuned the pre-trained models on the SceneFlow dataset~\cite{sceneflow} for 500 epochs using a mixed training set comprising KITTI 2012~\cite{kitti2012} and KITTI2015~\cite{kitti2015} training datasets. Batch size is set to 2. OneCycleLR scheduling is used with a max learning rate of 0.0002.
For the generalization experiments, we employ data augmentation techniques including color jitter, random erase, random scale, and random crop.
In the ablation study, these LightStereo models are trained over 50 epochs.

\begin{figure*}[t]
      \centering
           \includegraphics[width=0.95\linewidth]{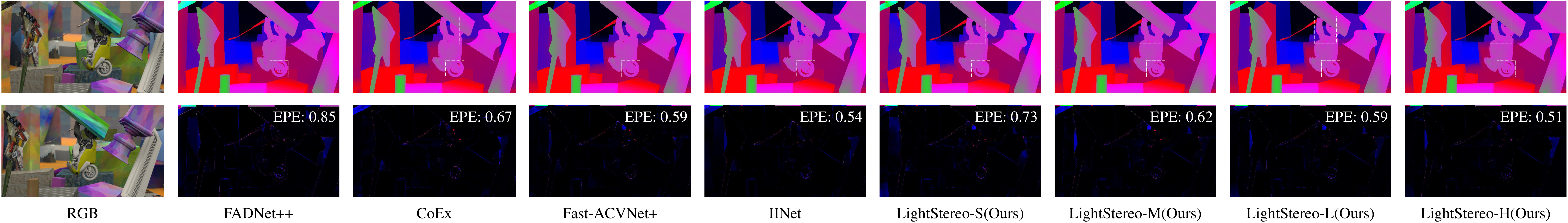}
      \caption{\textbf{Qualitative results on SceneFlow~\cite{sceneflow}.}} 
      \vspace{-0.38cm}
      \label{fig:visdifferent}
\end{figure*}


\begin{table}[t]
    \centering
    \caption{\textbf{Ablation study on SceneFlow~\cite{sceneflow} -- Conv. Block Selection.} 
    We \underline{underline} our final choice.
    }
    \resizebox{0.48\textwidth}{!}{
    \begin{tabular}{ccc|cccc}
     \toprule
     Conv. Type & Kernel&Block &EPE (px)&Flops(G)&Param(M)&Time(ms)  \\
     
     \midrule
     Regular & $3\times3$ & (4 8 16)&0.7652& 36.27&8.04&16.70\\
     Regular & $5\times5$ & (4 8 16) &0.7979&71.16&18.62&17.32\\
     Regular & $7\times7$ & (4 8 16) &0.8190&123.51&34.49&19.68\\
     Regular & $11\times11$ & (4 8 16) &0.8672&280.53&82.10&33.60\\
     \midrule
     V1 Block~\cite{mijwil2023mobilenetv1} & DW $3\times3$ &(30 60 120)&0.7801 &34.86&7.57&54.21\\ 
     \underline{V2 Block}~\cite{sandler2018mobilenetv2} & DW $3\times3$ & (4 8 16) &0.7144&35.82&7.54&22.93\\

     ViT Block~\cite{liu2023efficientvit}& - & (3 6 9)&0.7149 &34.48&6.53&51.14\\
     
    \bottomrule
    \end{tabular}
    }
\label{tab:ablationconvtype}
\vspace{-0.35cm}
\end{table}

\begin{table}[t]
    \centering
    \caption{\textbf{Ablation study on SceneFlow~\cite{sceneflow} -- Backbone selection.} Comparison among efficient feature extractors. }
    \resizebox{0.48\textwidth}{!}{
    \begin{tabular}{cc|cccc}
     \toprule
     Backbone &Type &EPE (px)&Flops(G)&Param(M)&Time(ms)  \\
     \midrule     
     MobilenetV2~\cite{sandler2018mobilenetv2}&CNN &0.7144&35.82&7.54&22.93\\
     MobilenetV3~\cite{mobilenetv3}&CNN&0.7292&35.72&9.16&25.02\\
     StarNet~\cite{StarNet} &CNN&0.7247&40.21&8.98&26.63\\
     EfficientnetV2 \cite{tan2021efficientnetv2}&CNN&0.6130&103.14&28.87&46.83\\
     RepVIT~\cite{wang2023repvit}&Transformer&0.6823&50.45&9.56&28.65\\
     
    \bottomrule
    \end{tabular}
    }
\label{tab:ablationbackbone}
\vspace{-0.25cm}
\end{table}

\begin{figure}[t]
      \centering
           \includegraphics[width=0.98\linewidth]{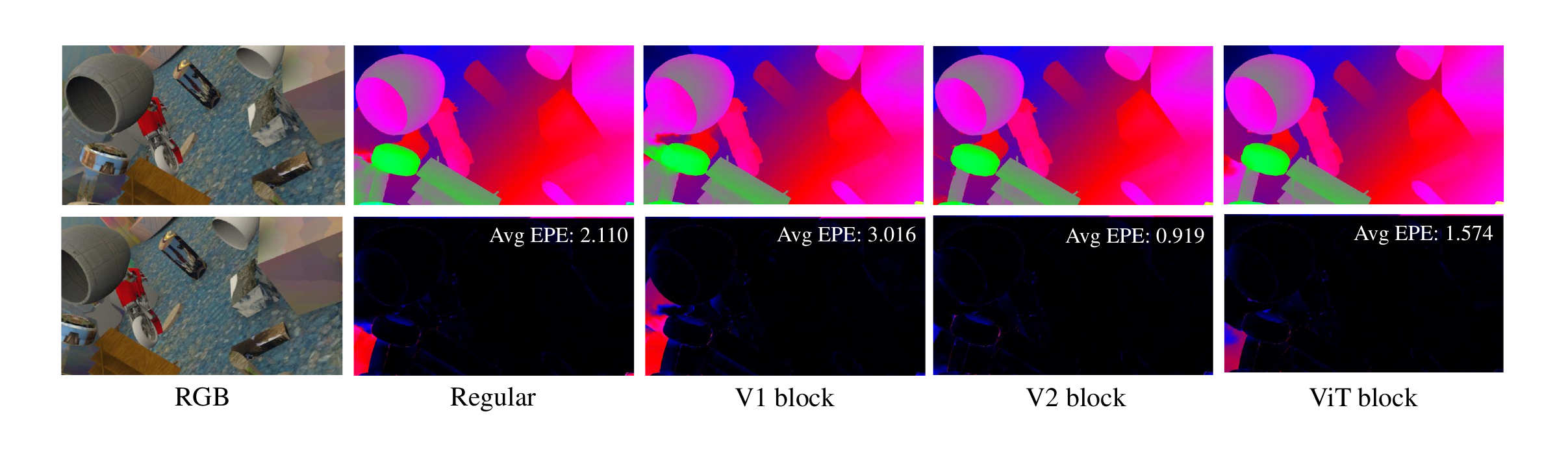}
       \vspace{-0.5cm}
      \caption{\textbf{Qualitative results on SceneFlow~\cite{sceneflow}.} Comparison between LightStereo blocks.} 
      \vspace{-0.38cm}
      \label{fig:visualOnSF}
\end{figure}

\begin{table}[t]
    \centering
    \caption{\textbf{Ablation study on SceneFlow~\cite{sceneflow} -- Block structure design.} The last row refers to LightStereo-L. 
    }
    \resizebox{0.48\textwidth}{!}{
    \begin{tabular}{c|cccc|cccc}
     \toprule
     & Block &Exp. factor &SE~\cite{hu2018squeeze}&MSCA &EPE (px)&Flops(G)&Param(M)&Time(ms)  \\
     \midrule
     a & (4 8 16)&(2 2 2)&&&0.7557&26.23&4.81&22.44\\
     b & (4 8 16)&(4 4 4)&&&0.7144&35.82&7.54&22.93\\
     c & (4 8 16)&(8 8 8)&&&0.6853& 54.99&12.99&23.59\\
     d & (4 8 16)&(16 16 16)&&&0.6650& 93.34&23.90&36.67\\
     
      \midrule
    e & (1 2 4)&(4 4 4)&&&0.8317& 22.17&3.34&16.65\\
    f & (2 4 8)&(4 4 4)&&&0.7464& 26.72&4.74&20.04\\
    g & (4 8 16)&(4 4 4)&&&0.7144&35.82&7.54&22.93\\
    h & (8 16 32)&(4 4 4)&&&0.6973 & 54.02&13.14&31.09\\
     
    \midrule
    i & (4 8 16)&(4 4 4)&&&0.7144&35.82&7.54&22.93 \\
    j & (4 8 16)&(4 4 4)& \checkmark&&0.7036 & 35.90&12.76&30.14\\
    k & (4 8 16)&(4 4 4) &&\checkmark&0.6809&36.36&7.64&23.14\\
    l & (4 8 16)&(4 4 4) &\checkmark&\checkmark& 0.6810&36.44&12.86&30.82\\
     \midrule
    m & (1 2 4) & (4 4 4) & &\checkmark&0.7899&22.71&3.44&17.59 \\
    n & (4 8 16)&(4 4 4) &&\checkmark& 0.6809&36.36&7.64&23.14\\
    o & (8 16 32)&(8 8 8)&&\checkmark&0.6382&91.85&24.29&37.55\\
    \bottomrule
    \end{tabular}
    }
\label{tab:ablationblock}
\vspace{-0.5cm}
\end{table}

\begin{table}[t]
    \centering
    \caption{
    \textbf{Results on KITTI online benchmarks.} Methods are classified based on runtime (higher or lower than 100ms). We highlight \textbf{best} and \underline{second} best results. $^*$ means time measured on our RTX 3090 GPU.}
    \renewcommand{\tabcolsep}{14pt}
    \scalebox{0.62}{
    \begin{tabular}{l|cc|ccc|c}
    \toprule
     \multirow{2}{*}{Method} & \multicolumn{2}{c|}{KITTI 2012\cite{kitti2012} } & \multicolumn{3}{c|}{ KITTI 2015\cite{kitti2015}}  & {Time}  \\
 & 3-all & EPE-all & D1-bg & D1-fg & D1-all& (ms)\\
 
    \midrule
    
 {GANet\cite{ganet2019}}  & 1.60 & 0.91 & 1.48 & 3.46 & 1.81  & 1800\\ 
 LaC+GANet\cite{LaC+GANet} & 1.42 &0.80 &1.44 & 2.83 & {1.67} &1800 \\ 
 CFNet\cite{cfnet2021} & 1.58 & 0.92 &1.54 &3.56 &1.88 &180 \\
 SegStereo\cite{segstereo2018} & 2.03 &1.25 &1.88 &4.07 &2.25 &600\\
SSPCVNet \cite{SSPCVNet} & 1.90 &1.08 &1.75 &3.89 &2.11 &900 \\
 EdgeStereo-V2\cite{song2020edgestereo} &1.83 &1.07 & 1.84 &3.30 &2.08 &320 \\
 LEAStereo\cite{leastereo} & {1.45} & {0.83} & 1.40 &{2.91} &1.65 &300\\
 CREStereo\cite{Crestereo} &1.46 &0.90 & 1.45 &2.86 &1.69 &410\\
 ACVNet~\cite{acvnet} & 1.47 & 0.86 &1.37 &3.07 &1.65 &200 \\
    \midrule
 {DispNetC\cite{sceneflow} } & 4.11 & 2.77 & 4.32 & 4.41 & 4.34 & 60\\
 {StereoNet \cite{stereonet2018}} & - & - & 4.30 & 7.45 & 4.83 & \underline{22}$^*$ \\
  {DeepPrunerFast\cite{deeppruner2019}} & - & - & 2.32 & 3.91 & 2.59 & 50\\
  {AANet\cite{xu2020aanet}} & 2.42  & 1.46 & 1.99 & 5.39 & 2.55 & 62\\
 DecNet\cite{DecNet}  & - & - & 2.07 & 3.87 & 2.37 & 50\\
  {HITNet\cite{hitnet}}  & \underline{1.41} & 1.14 & 1.74 & 3.20  & 1.98 & 54\\
  {CoEx\cite{bangunharcana2021coex}} & 1.55 & 1.15 & 1.79 & 3.82  & 2.13 & 33\\
  {Fast-ACVNet\cite{fast-acv}} &1.68&1.23&1.82&3.93&2.17& 39\\
  {Fast-ACVNet+\cite{fast-acv}} &1.45 & \underline{1.06} & \underline{1.70} & 3.53  & 2.01 & 45\\
 \textbf{LightStereo-S (Ours)} &1.88&1.30&2.00&3.80&2.30&\textbf{17$^*$ } \\
 \textbf{LightStereo-M (Ours)} &1.56&1.10&1.81&3.22&2.04&23$^*$ \\
 \textbf{LightStereo-L (Ours)} &1.55&1.10&1.78&\textbf{2.64}&\underline{1.93}&34$^*$  \\
    \textbf{LightStereo-H (Ours)} &\textbf{1.34}&\textbf{0.96}&\textbf{1.60}&\underline{2.92}&\textbf{1.82}&49$^*$\\
    \bottomrule
    \end{tabular}
    }
    \label{tab:evaluation_kitti}
    \vspace{-0.3cm}
\end{table}

\begin{table}[t]
\centering
\caption{\textbf{Runtime breakdown.}}
\renewcommand{\tabcolsep}{10pt}
\scalebox{0.7}{
\begin{tabular}{lccccc}
\toprule
Module & \makecell{Feature \\ Extraction}  & Cost  & \makecell{ Cost \\ Aggregation}& \makecell{Disparity \\ Regression} &\makecell{ Total \\ Time}\\ 
\midrule
LightStereo-S &10.39&1.98&3.98&1.48&17.83\\
LightStereo-M &10.39&1.98&9.59&1.49&23.45\\
LightStereo-L &10.39&1.98&23.64&1.49&37.50\\
LightStereo-H &27.17&1.98&23.64&1.49&54.28\\
\bottomrule
\end{tabular}
}
\label{tab:runtime}
\vspace{-0.5cm}
\end{table}

\begin{table}[t]
\centering
\caption{\textbf{Generalization performance on multiple datasets.} All the models are trained on the SceneFlow only.}
\renewcommand{\tabcolsep}{20pt}
\scalebox{0.7}{
\begin{tabular}{l|ccc}
\toprule
Method & \makecell{KITTI12 \\ D1(\%)}  & \makecell{KITTI15 \\ D1(\%)}  & \makecell{ Middle \\ 2px(\%)}\\
\midrule
DPF~\cite{deeppruner2019} &16.8&15.9&30.83\\
BGNet~\cite{bgnet} &24.8&20.1&37.00\\
CoEx~\cite{bangunharcana2021coex} &13.5&11.6&25.51\\
FastACV~\cite{fast-acv} &12.4&10.6&20.13\\
IINet~\cite{li2024iinet} &11.6&8.5&19.57\\
\textbf{LightStereo-S(Ours)} &11.6&9.0&19.63\\
\textbf{LightStereo-M(Ours)} &7.0&6.6&17.69\\
\textbf{LightStereo-L(Ours)} &\textbf{6.4}&\textbf{6.4}&17.51\\
\textbf{LightStereo-H(Ours)} &7.2&7.3&\textbf{14.27}\\
\bottomrule
\end{tabular}
}\vspace{-0.5cm}
\label{tab:generalization}
\end{table}


\subsection{Comparisons with State-of-the-art on SceneFlow}
Table \ref{tab:scene_flow} compares LightStereo with several state-of-the-art stereo matching approaches on SceneFlow~\cite{sceneflow}. 
LightStereo-S runs in 17ms only, being substantially faster than other methods. Regarding model complexity, LightStereo-S strikes a favorable balance with only 22.71Gflops, comparable to StereoNet~\cite{stereonet2018} (85.93Gflops). This reflects our commitment to maintaining a lightweight model while ensuring competitive performance. In terms of accuracy, LightStereo-H achieves an EPE of 0.51, resulting in more accuracy than Fast-ACVNet+~\cite{fast-acv} (0.59) and IINet~\cite{li2024iinet}(0.54), yet remaining competitive in terms of complexity.
Overall, our LightStereo framework, particularly the LightStereo-S configuration, presents a compelling solution for real-time stereo matching, offering a favorable trade-off between computational efficiency and depth estimation accuracy. 
Figure~\ref{fig:visdifferent} presents a qualitative comparison of the results by the four proposed models.

\subsection{Ablation Study}

\textbf{Conv. Block Selection.} 
In Table~\ref{tab:ablationconvtype}, we highlight the critical importance of disparity dimensions in the cost aggregation process of stereo matching, rather than spatial expansions in the height and width dimensions. On top, we can notice how regular convolutions with larger kernels yield both higher EPE and complexity.
This finding indicates that merely increasing the spatial extent of convolutions does not effectively enhance the accuracy of stereo-matching; on the contrary, putting the focus on disparity dimensions is crucial. 
The V1 block with a $3\times3$ kernel showed an EPE of 0.7801, indicating moderate performance with reduced FLOPs (34.86G), but it had a higher inference time (54.21ms). In contrast, the V2 block provided the best balance, achieving the lowest EPE of 0.7144, and FLOPs of 35.82G, with an inference time of 22.93ms. This superior performance is attributed to the structure of the V2 block, which incorporates expansion convolutions in the disparity dimension.
The ViT block showed a similar EPE to the V2 block but had a lower parameter count (6.53M) and higher inference time (51.14ms). Ultimately, the V2 block was chosen for its optimal balance between accuracy and computational efficiency. 
Figure~\ref{fig:visualOnSF} shows a visual comparison with different conv. type.  It can be observed that for the occluded area in the lower left corner of the left image, the cost aggregation based on the V2 block achieves better results than the other 3 conv. type.

\textbf{Backbone Selection.} 
Table~\ref{tab:ablationbackbone} explores the use of classic, lightweight models as backbones. MobilenetV2~\cite{sandler2018mobilenetv2} demonstrates a balanced performance with an EPE of 0.7144, FLOPs of 35.82G, 7.54M parameters, and an inference time of 22.93ms. Although MobilenetV3~\cite{mobilenetv3} had slightly lower FLOPs at 35.72G, it showed a higher EPE of 0.7292 and required more parameters (9.16M) with a longer inference time (25.02ms). 
StarNet~\cite{StarNet} exhibited an EPE of 0.7247 with higher FLOPs (40.21G) and inference time (26.63ms). EfficientnetV2~\cite{tan2021efficientnetv2} achieved the lowest EPE of 0.6130, but at the cost of significantly higher computational resources (103.14G FLOPs) and parameters (28.87M), with an inference time of 46.83ms. RepVIT~\cite{wang2023repvit}, a transformer-based model, showed an EPE of 0.6876 but required 50.45GFLOPs and 9.56M parameters with an inference time of 28.65ms. 
Therefore, MobilenetV2 is chosen for its overall efficiency and balance between accuracy and computational cost.

\textbf{Block Structure Analysis.} In Table~\ref{tab:ablationblock}, we explore the impact of different block structures while maintaining a constant expansion factor of 4. Configurations 'e' to 'h' explore blocks (1, 2, 4), (2, 4, 8), (4, 8, 16), and (8, 16, 32) respectively. The results show that larger blocks generally lead to better performance. For instance, configuration 'e' with the smallest block (1, 2, 4) has the highest EPE of 0.8317, while configuration 'h' with the largest block (8, 16, 32) improves EPE, but detailed metrics are not provided. There is a trade-off between accuracy and computational cost, as larger blocks increase FLOPs and parameters.

\textbf{Expansion Factor Analysis.} Still in Table~\ref{tab:ablationblock}, we analyze the effect of varying expansion factors while keeping the block structure constant.
As the expansion factor increases, a consistent decrease in EPE is observed, indicating improved accuracy.
The analysis further proves that the critical importance of disparity dimensions cannot be overstated in the cost aggregation process of stereo matching. The disparity dimension is pivotal in effectively aggregating cost volumes, leading to more accurate and reliable depth estimations.
However, this improvement in accuracy comes at the cost of significantly increased computational complexity and parameters, with flops increasing from 26.23G to 93.34G and the inference time increases from 22.44ms to 36.67ms.

\textbf{MSCA Module Analysis.} As illustrated in Table~\ref{tab:ablationblock}, the baseline configuration (i) with blocks (4, 8, 16) achieved an EPE of 0.7144. Incorporating MSCA (configuration k) reduced the EPE further to 0.6809 with minimal changes in FLOPs and parameters. This suggests that MSCA provides the most significant improvement in accuracy with minimal impact on computational efficiency.
We also explored the use of the SE module~\cite{hu2018squeeze}. Configurations 'i' to 'j' show that although the SE module improves accuracy, reducing EPE from 0.7144 to 0.7036.
However, this improvement comes with an increase in parameters from 7.54M to 12.76M and a slight increase in FLOPs and inference time.

\textbf{Runtime Analysis.} In, Table~\ref{tab:runtime}, LightStereo demonstrates commendable efficiency across its constituent modules. 
As the number of blocks increases, the time required for cost aggregation in LightStereo-S, LightStereo-M, and LightStereo-L also increases. For LightStereo-H, the time spent on feature extraction is extended due to the adoption of a more complex EfficientNetV2.
These runtime breakdowns underscore the effectiveness of the LightStereo framework in achieving real-time stereo-matching capabilities.

\subsection{Comparison with State-of-the-art on Real Datasets}

\textbf{KITTI benchmarks.}
We evaluate the results of our LightStereo variants on the KITTI 2012 and 2015 online benchmarks. As illustrated in Table~\ref{tab:evaluation_kitti}, LightStereo-S runs the fastest. Notably, LightStereo-H surpasses all other lightweight state-of-the-art stereo-matching networks across every evaluation metric on KITTI 2012. Additionally, LightStereo-H achieved the best results in terms of D1-bg and D1-all among lightweight models on KITTI 2015.

\textbf{Cross-domain generalization.} We evaluated the generalization performance of our model on real-world datasets such as KITTI12~\cite{kitti2012}, KITTI15~\cite{kitti2015}, Middlebury~\cite{middlebury}, 
as shown in Table~\ref{tab:generalization}. 
LightStereo shows superior generalization over other lightweight methods, which indicates that our approach has moderate complexity to avoid overfitting.

\section{Conclusion} 

This paper designs a lightweight 2D encoder-decoder aggregation network to achieve precise and fast disparity estimation, called LightStereo.
While similar approaches have been explored previously, our novel contribution lies in optimizing performance through a targeted emphasis on the disparity channel dimension within the 3D cost volume, which encapsulates the distribution of matching costs. Our exhaustive exploration has led to the development of numerous strategies to enhance the capacity of this crucial dimension, ensuring both precision and efficiency in disparity estimation. LightStereo offers a compelling solution for accelerating the matching process while maintaining high levels of accuracy and efficiency.

\textbf{Acknowledgements.} This work was supported by the National Natural Science Foundation of China under Grant 62373356 and the Joint Funds of the National Natural Science Foundation of China under U24B20162.

\bibliographystyle{IEEEtran}
\bibliography{main}

\end{document}